\documentclass[letterpaper, 10 pt, conference]{ieeeconf}  %

\IEEEoverridecommandlockouts                              %

\usepackage{amsmath} %
\usepackage{amssymb}  %
\usepackage{booktabs}  %
\usepackage{listings}  %
\usepackage{graphicx}  %
\usepackage{mwe}
\usepackage[utf8]{inputenc}
\usepackage{multirow}  %
\usepackage{caption}
\usepackage{makecell}
\usepackage{siunitx}

\usepackage[
            url=false,
            isbn=false,
            doi=false,
            backref=false,
            style=ieee,
            mincitenames=1,
            maxcitenames=1,
            citestyle=numeric-comp,
            sorting=none,
            block=none]{biblatex}
\renewcommand{\bibfont}{\small}
\newcommand{\algname}{\texttt{PyRoki}}
\usepackage{etoolbox}

\usepackage{hyperref}

\usepackage{xcolor}
\definecolor{mplblue}{RGB}{31, 119, 180}
\definecolor{mplorange}{RGB}{255, 127, 14}
\usepackage{pifont}
\definecolor{darkgreen}{RGB}{0,120,0}
\definecolor{darkred}{RGB}{180,0,0}
\newcommand{\cmark}{\textcolor{darkgreen}{\ding{51}}}

\newcommand{\xmark}{\textcolor{darkred}{\ding{55}}}

\usepackage{tikz}
\usepackage{pgfplots}
\pgfplotsset{compat=1.18}
\usepackage{pgfplotstable}

\title{\LARGE \bf
\algname{}: A Modular Toolkit for Robot Kinematic Optimization
}
\author{
    Chung Min Kim$^{*}$ \and
    Brent Yi$^{*}$ \and
    Hongsuk Choi \and
    Yi Ma\and
    Ken Goldberg \and
    Angjoo Kanazawa\\
}

\begin{document}

\twocolumn[{%
    \renewcommand\twocolumn[1][]{#1}%
    \maketitle
    \centering
    
    \vspace{-1.5em}
    UC Berkeley\\
    
    \vspace{1.5em}
    \url{https://pyroki-toolkit.github.io} \\
    
    \vspace{1.5em}
    
    \includegraphics[width=\textwidth,trim=3 3 3 10,clip]{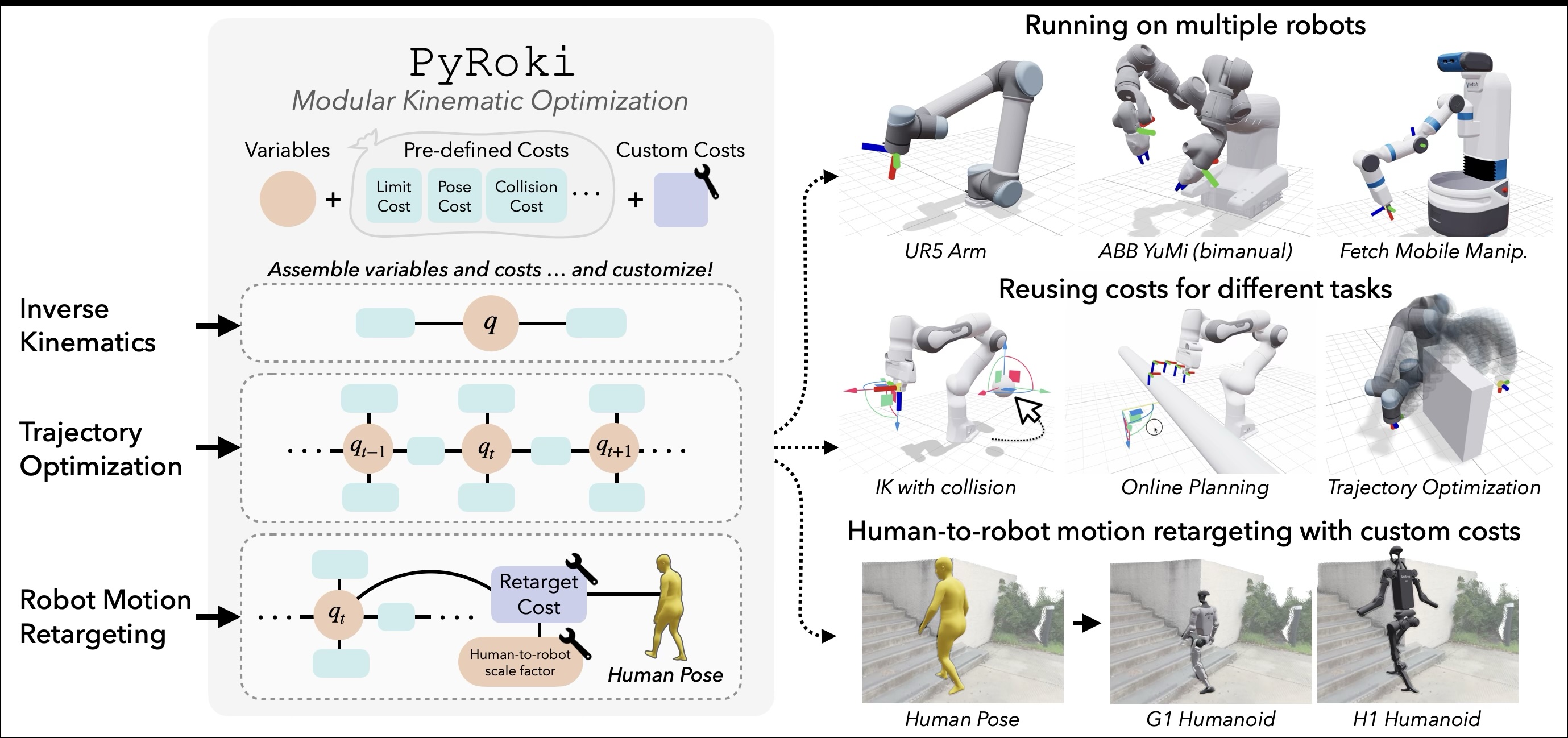}
    \captionof{figure}{
        \textbf{\algname{} is a modular, extensible, and cross-platform toolkit for kinematic optimization.}
        We unify problems like inverse kinematics, trajectory optimization, and motion retargeting using composable kinematic variables and costs.
        \algname{} aims to support a broad variety of robots and tasks, and runs on CPU, GPU, and TPU.
    }
    \label{fig:splash}
}]

\thispagestyle{empty}
\pagestyle{empty}

\begin{abstract}
Robot motion can have many goals.
Depending on the task, we might optimize for pose error, speed, collision, or similarity to a human demonstration.
Motivated by this, we present \algname{}: a modular, extensible, and cross-platform toolkit for solving kinematic optimization problems. 
\algname{} couples an interface for specifying kinematic variables and costs with an efficient nonlinear least squares optimizer. %
Unlike existing tools, it is also cross-platform: optimization runs natively on CPU, GPU, and TPU.
In this paper, we present (i) the design and implementation of \algname{}, (ii) motion retargeting and planning case studies that highlight the advantages of \algname{}'s modularity, and (iii) optimization benchmarking, where \algname{} can be 1.4-1.7x faster and converges to lower errors than cuRobo, an existing GPU-accelerated inverse kinematics library.
\let\thefootnote\relax\footnotetext{$^*$Equal contribution.}

\end{abstract}

\begin{table*}[t]
    \centering
    \renewcommand{\arraystretch}{1.1}
\begin{tabular}{@{}l*{3}{c}*{3}{c}*{3}{c}*{2}{c}@{}}
    \toprule
    & \multicolumn{3}{c}{\textbf{Hardware}} & \multicolumn{3}{c}{\textbf{Supported Robots}} & \multicolumn{3}{c}{\textbf{Tasks}} & \multicolumn{2}{c}{\textbf{Features}} \\
    \cmidrule(lr){2-4} \cmidrule(lr){5-7} \cmidrule(lr){8-10} \cmidrule(l){11-12}
    \textbf{Method} & CPU & GPU & TPU & Arm & Hand & Humanoid & 
    \makecell{IK} & \makecell{Traj.\\Opt.} & \makecell{Retargeting} & 
    \makecell{Collision} & \makecell{Custom\\Costs} \\
    \midrule
    TracIK~\cite{beeson2015trac} &
        \cmark & \xmark & \xmark & 
        \cmark & \cmark & \cmark &
        \cmark & \xmark & \xmark & 
        \cmark & \xmark \\ %
    pink~\cite{pink}, mink~\cite{Zakka_Mink_Python_inverse_2024} &
        \cmark & \xmark & \xmark & 
        \cmark & \cmark & \cmark &
        \cmark & \xmark & \xmark & 
        \cmark & \xmark \\ %
    TrajOpt~\cite{schulman2014motion} &
        \cmark & \xmark & \xmark & 
        \cmark & \cmark & \cmark &
        \xmark & \cmark & \xmark & 
        \cmark & \xmark \\ %
    cuRobo~\cite{curobo_icra23} & 
        \xmark & \cmark & \xmark & 
        \cmark & \xmark & \xmark &
        \cmark & \cmark & \xmark & 
        \cmark & \cmark \\ %
    Dex-Retargeting~\cite{qin2023anyteleop} & 
        \cmark & \xmark & \xmark & 
        \xmark & \cmark & \xmark &
        \cmark & \xmark & \cmark & 
        \xmark & \xmark \\ %
    H2O~\cite{he2024learning} & 
        \cmark & \xmark & \xmark & 
        \xmark & \xmark & \cmark &
        \cmark & \xmark & \cmark & 
        \xmark & \xmark \\ %
    \cmidrule(r){1-1} \cmidrule(lr){2-4} \cmidrule(lr){5-7} \cmidrule(lr){8-10} \cmidrule(l){11-12}
    \algname{} & 
        \cmark & \cmark & \cmark & 
        \cmark & \cmark & \cmark &
        \cmark & \cmark & \cmark & 
        \cmark & \cmark \\ %
    \bottomrule
\end{tabular}
\vspace{0.5em}
    \caption{\textbf{Comparison between \algname{} and a selection of existing kinematic optimization tools.} This table summarizes the documented features and supported functionalities of different frameworks, highlighting the broad capabilities of \algname{}.
    }
    \vspace{-0.5em}
    \label{tab:framework_comparison}
\end{table*}

\section{Introduction}
Numerical optimization is the standard solution for many tasks in robot kinematics.
Using objectives like pose error~\cite{strub2020adaptively}, smoothness~\cite{rakita2018relaxedik}, and similarity to a human demonstration~\cite{handa2020dexpilot,qin2023anyteleop} the robotics community has built diverse optimization software for tasks such as inverse kinematics (IK)~\cite{beeson2015trac,wang2023rangedik,Zakka_Mink_Python_inverse_2024,Zhong_PyTorch_Kinematics_2024,meier2022differentiable}, trajectory optimization~\cite{schulman2014motion,ratliff2009chomp,bhardwaj2022storm,curobo_icra23,kalakrishnan2011stomp,chitta2012moveit,sucan2012open,kunz2012time}, and motion retargeting~\cite{he2024learning,handa2020dexpilot,qin2023anyteleop}.
These tools are fast, mature, and widely adopted in both research and production.
Despite the shared structure of kinematic optimization, existing tools are fragmented. %
Implementations typically rely on task-specific C++ routines~\cite{beeson2015trac}, CUDA kernels~\cite{curobo_icra23}, or task-specific analytical Jacobians~\cite{Zakka_Mink_Python_inverse_2024}.
While these features can improve efficiency, they create barriers for incorporating new objectives.
As shown in Table~\ref{tab:framework_comparison}, different task and robot hardware variations can require different tools.
This fragmentation prevents tools from transferring between related problems.
Low-level specialization also constrains computation: optimizers tend to be restricted to either only CPU or only GPU operation.
We present \algname{} (\textbf{Py}thon \textbf{Ro}bot \textbf{Ki}nematics), a modular, extensible, and cross-platform
toolkit for kinematic optimization.
The core idea behind \algname{} is that many tasks that currently require disjoint tools---such as IK, trajectory optimization, and motion retargeting---solve similar optimization problems.
\algname{} provides (i) an interface for specifying these problems using modular variable and cost function abstractions, (ii) the ability to efficiently solve them using a Levenberg-Marquardt optimizer, and (iii) a web-based visualizer for interactively tuning cost weights.
Furthermore, \algname{} is designed for efficient execution on CPU, GPU and TPU.
The goal of \algname{} is to streamline kinematic optimization for robotics, just as deep learning frameworks like PyTorch~\cite{paszke2019pytorch} have made it easier to define and experiment with deep learning models.

The contributions of this paper are as follows:
\begin{enumerate}
    \item \textbf{Toolkit.} We present \algname{}, a modular toolkit for kinematic optimization for robotics.
    \algname{} directly supports standards like URDF~\cite{quigley2009ros}, and can run on CPU, GPU, and TPU. It also includes a real-time web-based visualizer to explore robot behavior, such as adjusting the impact of cost functions and adding new scene visualizations. %

    \item \textbf{Applications.} We demonstrate the effectiveness of \algname{}'s unified framework across multiple tasks, including inverse kinematics, trajectory optimization, and motion retargeting for robot hands and humanoids.

    \item \textbf{Benchmarking.}
    \algname{} solves optimization problems using a Levenberg-Marquardt solver. We evaluate \algname{} using both automatic and analytic Jacobians. For batched IK, \algname{} can be 1.4-1.7x faster than previous GPU-accelerated methods~\cite{curobo_icra23}.

\end{enumerate}

\noindent All code is released with an open-source license.

\section{Related work} \label{sec:related_work}

\subsection{Inverse Kinematics}

The goal of inverse kinematics (IK) is to recover a joint configuration that achieves a desired end-effector pose.
Classical approaches focus on recovering analytical, closed-form solutions for this problem~\cite{diankov_thesis,paden1985kinematics,pieper1969kinematics,paul1981robot,raghavan1990kinematic,husty1996algorithm}.
These approaches are efficient, but require assumptions on kinematic configuration.
Recent approaches have therefore focused more on iterative optimization~\cite{beeson2015trac,bruyninckx-icra2003,curobo_icra23,wang2023rangedik,rakita2021collisionik,Rakita-RSS-18,Zakka_Mink_Python_inverse_2024,whitney1969resolved,whitney1972mathematics,rakita2018relaxedik}, which is more flexible.
Optimization enables secondary objectives for controlling redundant degrees of freedoms~\cite{liegeois1977automatic}, both geometric~\cite{gleicher1998retargetting,Rakita-RSS-18,curobo_icra23} and learned~\cite{rakita2021collisionik} collision constraints, 
Critically, iterative optimization also generalizes across robot embodiments. %
The goal of \algname{} is to take this flexibility a step further.
Rather than specialize for the IK task alone, \algname{} provides a kinematic optimization toolkit where IK can be defined using a subset of possible cost functions. %

\subsection{Trajectory Optimization}

While inverse kinematics focuses on solving for a single joint configuration, the goal of trajectory optimization is to output a continuous sequence of joint configurations.
Methods include trajectory optimization with collision costs~\cite{ratliff2009chomp,bhardwaj2022storm,kalakrishnan2011stomp,schulman2014motion,mukadam2016gaussian,mukadam2018continuous}, applications with multiple goal sets~\cite{dragan2011manipulation}, and simultaneous grasp and trajectory optimization~\cite{wang2019manipulation,ichnowski2020gomp}.
An important insight made by~\cite{mukadam2018continuous} is that trajectory optimization can be accelerated by exploiting structure: each configuration in the trajectory interacts with its temporal neighbors, leading to sparse optimization problems.
Trajectory optimization can therefore benefit from sparse linear algebra routines~\cite{dellaert2017factor,yi2021differentiable,schoenberger2016sfm,Agarwal_Ceres_Solver_2022,kummerle2011g2o}, which \algname{} supports.

\subsection{Motion Retargeting}

Motion retargeting is the problem of transferring motion from a source embodiment to a target embodiment~\cite{gleicher1998retargetting,ho2010spatial} (e.g., from human to robot).
Retargeting is critical step for many systems for robot teleoperation~\cite{qin2023anyteleop,schmidt2014dart} and for learning from human demonstrations~\cite{peng2018sfv,he2024learning,fu2024humanplus,zhang2022kinematic}, but is made challenging by kinematic differences between embodiments---it requires balancing both similarity and physical plausibility objectives~\cite{zhang2023skinned,cheynel2023sparse}.
Different application contexts also impose distinct computational requirements: for teleoperation, motion retargeting might need to run in real-time on a single CPU to control a single robot's motion.
In contrast, batch processing on GPUs can dramatically accelerate offline processing for learning from large-scale datasets~\cite{AMASS:ICCV:2019,li2021egoexo,lin2023motionx,bollo2024nymeria}.
\algname{} aims to simplify optimization-based retargeting by making it easier to compose and tune costs, while introducing efficient parallelization on GPUs.

\subsection{Modular Optimization Tools}
Modular optimization tools have transformed numerous scientific and engineering disciplines.
Frameworks like TensorFlow~\cite{abadi2016tensorflow} and PyTorch~\cite{paszke2019pytorch} have revolutionized machine learning with automatic differentiation and hardware flexibility.
JuMP~\cite{Lubin2023Jump} provides a mathematical optimization framework with interchangeable solvers, while Drake~\cite{drake}, Ceres Solver~\cite{Agarwal_Ceres_Solver_2022}, g2o~\cite{kummerle2011g2o}, GTSAM~\cite{gtsam}, miniSAM~\cite{dong2019minisam}, CasADi~\cite{andersson2019casadi}, and ACADO~\cite{houska2011acado} offer specialized capabilities for robotics, computer vision, and control applications.
Inspired by these tools, \algname{} aims to simplify kinematic optimization for diverse robots, tasks, and compute platforms.
We summarize relevant features in Table~\ref{tab:framework_comparison}.

\section{\algname{}: Modular Kinematic Optimization}

We propose \algname{}, a robot kinematics library that is designed around three goals:

\subsubsection{\textbf{Modular}} 
\algname{} separates optimization variables from cost functions, creating reusable components that work across different tasks. This lets the same objectives (e.g., collision avoidance, pose matching) apply to multiple tasks (e.g., inverse kinematics or trajectory optimization) without reimplementation.

\subsubsection{\textbf{Extensible}}
\algname{} supports rapid prototyping through automatic differentiation~\cite{jax2018github}, computing Jacobians for user-defined cost functions, as well as a real-time interface for tuning cost weights. This simplifies custom optimization objectives, while also supporting analytical Jacobians when performance is critical.

\subsubsection{\textbf{Cross-Platform}} 
\algname{} runs natively on CPUs, GPUs, and TPUs, enabling seamless scaling from single-robot optimization to parallel batch processing. Cross-platform capabilities can accelerate demanding applications like processing for large motion datasets or sampling-based planning~\cite{curobo_icra23}. %

There are three components of \algname{} that make these goals possible: a quasi-Newton optimizer, variable abstractions, and composable cost functions.

\subsection{Solver Backbone} \label{sec:solver}

Kinematic optimization benefits from quasi-Newton approaches, which accelerate convergence using cost curvature approximations.
\algname{} uses a Levenberg-Marquardt (LM) optimizer~\cite{levenberg1944method,marquardt1963algorithm, wampler1986manipulator, nakamura1986inverse, buss2005selectively}.
Our optimizer builds on prior work~\cite{yi2021differentiable,yi2024estimating,heppert2022category} and uses JAX~\cite{jax2018github}, a high-level array programming interface that enables parallelization on CPU, GPU, and TPU.
For efficiency, it automatically computes block-sparse Jacobian matrices; this is particularly advantageous for temporally sparse motion planning problems~\cite{mukadam2018continuous}.
\algname{}'s optimizer does not directly handle hard constraints, but we outline in Section \ref{sec:costs} how we represent joint limits and contact avoidance with differentiable penalties.

\begin{figure}[t!]
    \centering
    \includegraphics[width=0.95\linewidth,trim=2 2 2 2,clip]{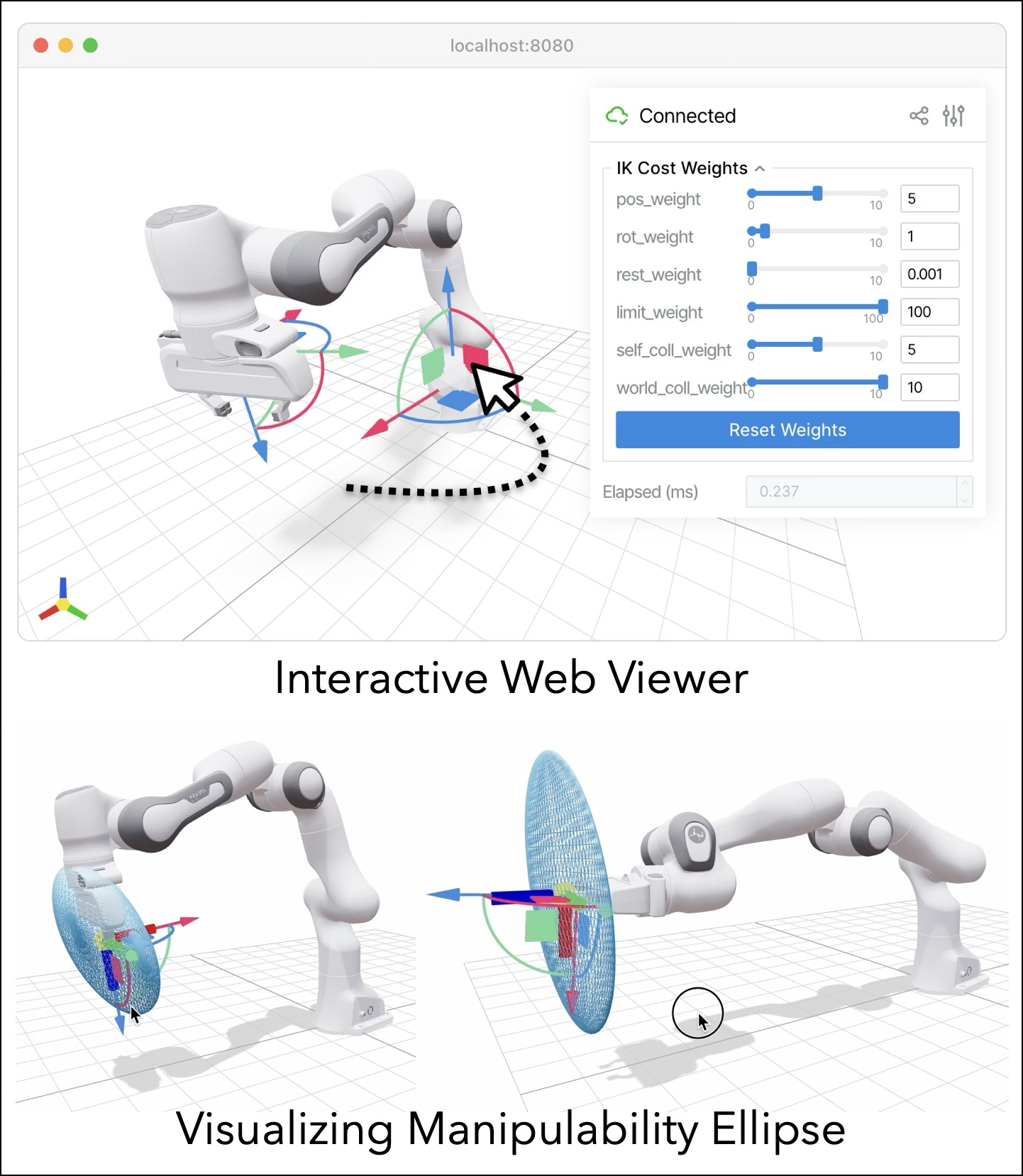}
    \caption{\textbf{Interactive Web-based Robot Viewer}.
    Users can tune weights for the different costs in real-time using a web interface (top), built on viser~\cite{viser}. The viewer can also display the robot's configuration, set a goal, or modify the environment. The user can also add additional visualization, e.g., manipulability ellipse (bottom).}
    \label{fig:interactive_robot_viewer}
    \vspace{-1em}
\end{figure}

\subsection{Variable Abstractions}

For modularity, \algname{} defines abstractions for common kinematic optimization variables. 
The core abstraction is the joint configuration variable, representing robot articulation states.
We represent joint configurations as $\mathbf{q}$, with timesteps denoted as $\mathbf{q}_{t}$. Forward kinematics maps these configurations to poses, with $\mathbf{T}_{base \leftarrow i} = FK(\mathbf{q})_{i}$ representing the transform between the robot base and joint $i$.
We support fixed, revolute, and prismatic joints, along with mimic joints commonly found in robot hands.

Kinematic variables can also be composed with SE(3) and SO(3) Lie group variables for representing poses and orientations. %
Operations like interpolation, composition, and pose error computation respect geometric structure without requiring users to implement complex manifold operations.

\subsection{Cost Functions} \label{sec:costs}

\algname{}'s modular design is based on the idea of composable cost functions.
Different objectives can be combined, weighted, and reused across different optimization tasks. %

In addition to supporting custom cost functions, \algname{} also comes with pre-implemented costs for common kinematic optimization tasks. 
We survey these in their residual forms below; in practice, these are also weighted.
The LM optimizer minimizes the sum of squared residuals.

\textbf{Joint Pose Cost.}
The joint pose cost $\hat{c}_{\text{pose}}$ penalizes the difference between current and target joint poses:
$$\hat{c}_{\text{pose}}(\mathbf{q}, i, \mathbf{T}_{\text{base} \leftarrow \text{target}}) = \log(\mathbf{T}_{\text{base} \leftarrow \text{target}}^{-1}  \mathbf{T}_{\text{base} \leftarrow i}) $$
This cost can be quickly modified---for example, to include the base pose of a mobile manipulator.
We evaluate this in Section~\ref{sec:ik_mobile_base}.

\textbf{Joint Limit Cost}
The joint limit cost $\hat{c}_{\text{limit}}$ penalizes joint values that are outside of their mechanical bounds:
$$\hat{c}_{\text{limit}}(\mathbf{q}) = \max(0, \mathbf{q} - \mathbf{q}_{\text{upper}}) + \max(0, \mathbf{q}_{\text{lower}} - \mathbf{q})$$
\noindent where $\mathbf{q}_{\text{upper}}$ and $\mathbf{q}_{\text{lower}}$ are the upper and lower mechanical limits of the joint, respectively.

\textbf{Velocity Limit Cost}
The velocity limit cost $\hat{c}_{\text{vel}}$ penalizes joint velocities that are outside of their mechanical limits:
$$\hat{c}_{\text{vel}}(\mathbf{q}, \dot{\mathbf{q}}) = \max(0, |\dot{\mathbf{q}}| - \dot{\mathbf{q}}_{\text{limit}} \cdot dt)$$
\noindent where $\dot{\mathbf{q}}_{i,\text{limit}}$ is the velocity limit of the joint, and $dt$ is the time step between configurations.

\textbf{Joint Regularization Cost.} The joint regularization cost $\hat{c}_{\text{reg}}$ encourages the solution to be close to a user-defined default pose, as redundant robots (e.g., 7-DoF arms) can reach multiple configurations. %
$$\hat{c}_{\text{reg}}(\mathbf{q}) = \mathbf{q} - \mathbf{q}_{\text{reg}}$$

\textbf{Smoothness Cost.} The smoothness cost $\hat{c}_{\text{smooth}}$ encourages small changes in joint positions, and is useful for generating smooth trajectories.
$$\hat{c}_{\text{smooth}}(\mathbf{q}, t) = \mathbf{q}_{t} - \mathbf{q}_{t-1}$$
Similar costs can be written for acceleration and jerk minimization by approximating from $\mathbf{q}_{t}$ with the five-point stensil method~\cite{curobo_icra23}.

\textbf{Manipulability Cost.}
The manipulability cost $\hat{c}_{\text{manip}}$ penalizes configurations where the robot is close to a singularity, and maximizes the Yoshikawa's manipulability measure~\cite{yoshikawa1985manipulability}.
$$\hat{c}_{\text{manip}}(\mathbf{q}, i) = \left(\sqrt{\det(\mathbf{J}_i(\mathbf{q})\mathbf{J}_i(\mathbf{q})^T)} + \epsilon\right)^{-1}$$
where $\mathbf{J}_i(\mathbf{q})$ is the manipulator Jacobian of the robot at configuration $\mathbf{q}$ for the $i$-th robot joint, and $\epsilon$ is a small constant to prevent division by zero.

\textbf{Collision Avoidance Costs.} 
Self and world collision costs are critical for generating feasible robot motion. Following prior work~\cite{curobo_icra23,wang2023rangedik}, we implement them as:

$$\hat{c}_{\text{self\_coll}}(\mathbf{q}) = \sum_{(i,j) \in \text{links}} f(d_{ij}(\mathbf{q}), \eta_{ij})$$

$$\hat{c}_{\text{world\_coll}}(\mathbf{q}) = \sum_{(i, j) \in \text{links,obs}} f(d_{ij}(\mathbf{q}), \eta_{ij})$$

\noindent where $d_{ij}$ is the signed distance between collision geometries for link pairs $(i,j)$ at joint configuration $\mathbf{q}$ ($d_{ij} < 0$ if in collision). %
The signed distance is then converted into a cost $d_c$ using a smooth activation function $f(d, \eta)=$ that avoids a discontinuity at $d = 0$~\cite{ratliff2009chomp,curobo_icra23} and penalizes 
\begin{equation}
d_c = \begin{cases}
    -d + 0.5\eta & \text{if } d < 0 \\
    \frac{0.5}{\eta}(-d + \eta)^2 & \text{if } 0 < d < \eta \\
    0 & \text{otherwise}
    \end{cases}
\end{equation}
where $\eta$ is the buffer distance.
In Section~\ref{sec:ik_trajopt}, we show how collision can also be considered across timesteps. %

\subsection{Interactive Robot Web Viewer} \label{sec:interactive_viewer}

Good visualization is critical for 3D robotics tasks. \algname{} includes an interactive web-based viewer built on viser~\cite{viser}, which provides an intuitive way to explore robot behavior in real time.
With the viewer, users can interactively adjust cost function weights and immediately see how they affect robot motion, helping to balance competing objectives like end-effector accuracy, joint limits, and collision avoidance.
Beyond weight tuning, the viewer allows interactive scene modifications (moving obstacles) and custom elements like manipulability ellipses (Figure~\ref{fig:interactive_robot_viewer}).

\begin{figure}
    \centering
    \includegraphics[width=0.9\linewidth,trim=2 2 2 2,clip]{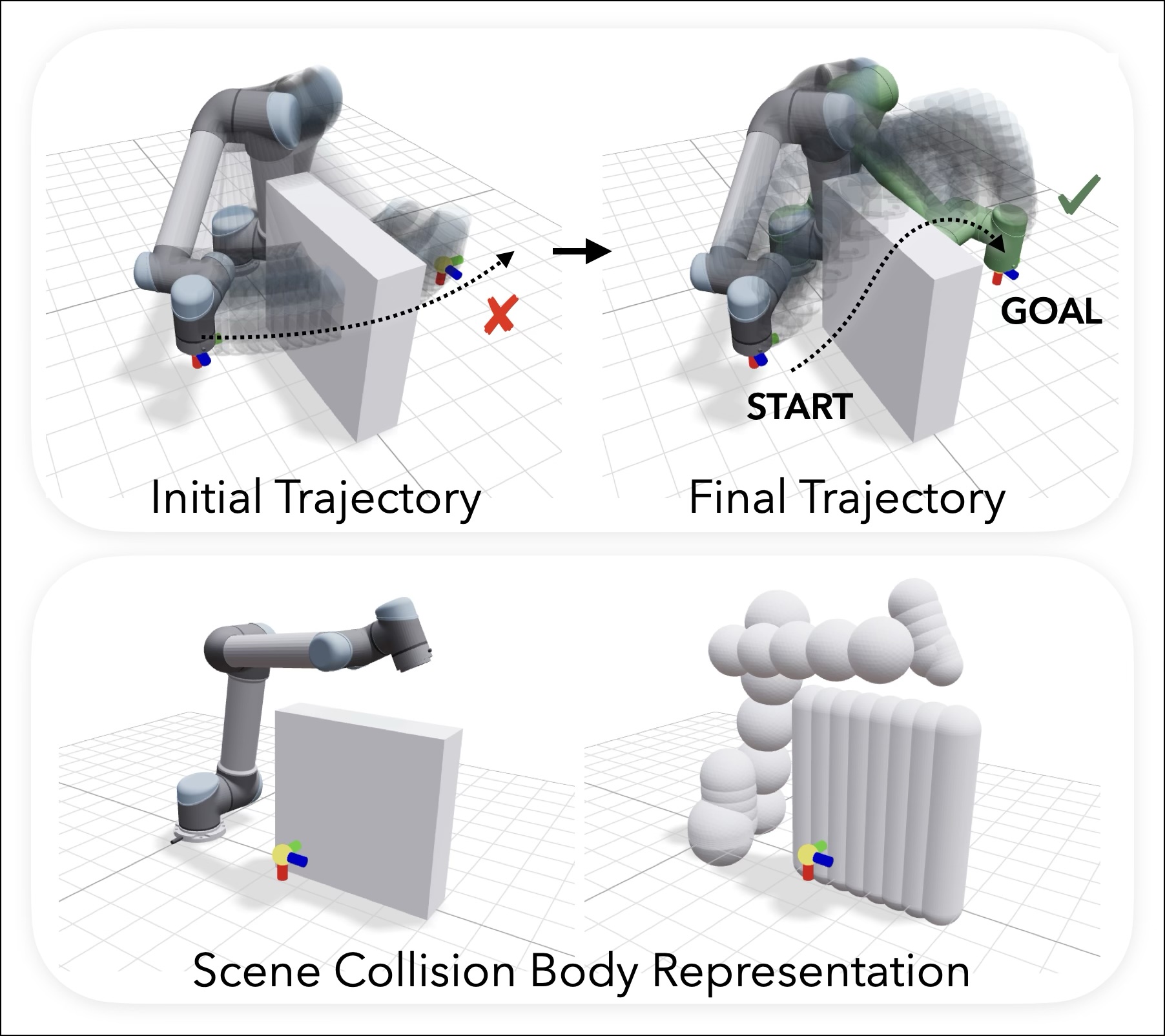}
    \caption{\textbf{Trajectory Optimization.} \algname{} can be used to formulate trajectory optimization problems that find valid collision-free solutions from naive straight-line initializations (top), similar to CHOMP~\cite{ratliff2009chomp}. The arm is approximated as spheres (bottom), which are connected into capsules for collision checking between neighboring timesteps. %
    }
    \label{fig:trajopt}
\end{figure}

\begin{figure*}[!ht]
    \centering
    \includegraphics[width=0.9\linewidth,trim=4 10 4 10,clip]{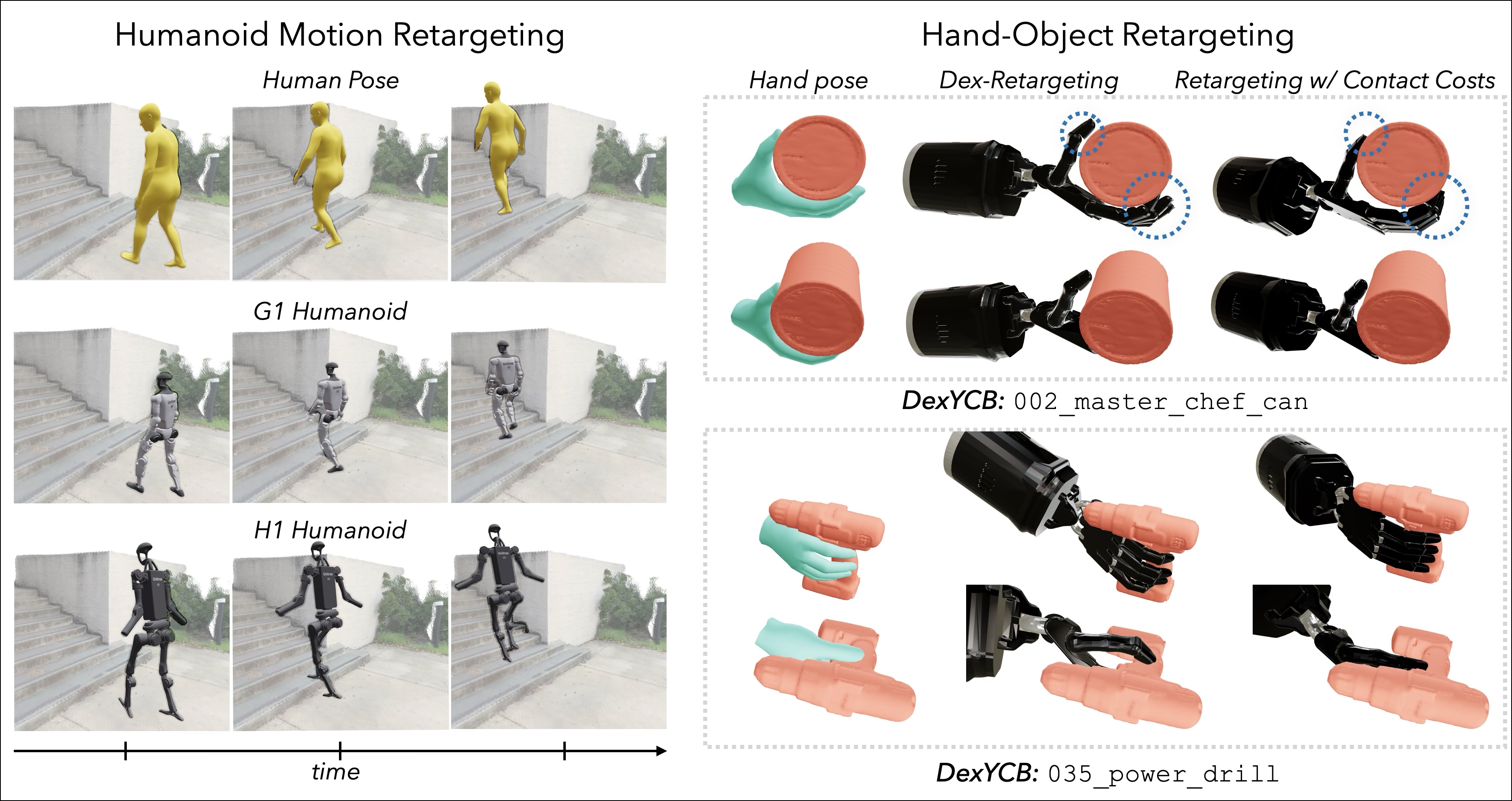}
    \caption{
    \textbf{Robot Motion Retargeting.} We show motion retargeting for humanoids and robot hands using \algname{}, using the \textit{same} motion transfer cost across robots and tasks. 
    To handle differences in robot morphology, we optimize for robot joint configurations and per-link scaling factors between embodiments simultaneously.
    Contact costs ensure humanoids stay grounded through scene contact (left) and maintain fingertip-object contact when present in the source motion (right). Blue dotted lines indicate contact relationships between the robot hand and the object.
    }
    \label{fig:retargeting_results}
    \vspace{-1.5em}
\end{figure*}

\section{Qualitative Results}
We illustrate \algname{}'s modularity and extensibility using four robot kinematic tasks, which use both pre-defined (Sec.~\ref{sec:costs}) and custom costs.
We begin with a case study on IK and trajectory optimization, drawing on existing optimization-based motion planning approaches~\cite{ratliff2009chomp,schulman2014motion}.
We then examine \algname{} for robot motion retargeting, transferring human motions to humanoids and robot hands.

\subsection{Inverse Kinematics and Trajectory Optimization} \label{sec:ik_trajopt}

Writing inverse kinematics in \algname{} is straightforward; the user only needs to import their robot URDF and assemble cost terms defined in Section~\ref{sec:costs}, such as the joint pose cost, joint limit costs, and joint regularization cost.
This formulation can be easily extended to find IK solutions close to the current state, or to find collision-free solutions.
Moreover, these tasks can be executed seamlessly across a diverse range of robotic platforms, as shown in Figure~\ref{fig:splash}.

Much of the logic for IK can be re-used for trajectory optimization. We show an example setting where a UR5 robot arm moves between specified start and goal poses while avoiding an obstacle in the middle, as shown in Figure~\ref{fig:trajopt}. The trajectory is optimized with the same costs in IK, but also including collision avoidance and acceleration- and jerk- minimization.
We solve for the initial trajectory by first determining collision-free start and goal joint configurations using inverse kinematics. We then linearly interpolate between these configurations to create a trajectory, which may initially be in collision with the environment.

We additionally implement continuous collision costs through sweeping volumes~\cite{schulman2014motion,ratliff2009chomp} to check collision between timesteps.
The UR5 robot is modeled as a series of spheres (see Fig.~\ref{fig:trajopt}, bottom), and each sphere associated with the UR5 at timestep $t$ is connected to the corresponding sphere at the next timestep to form a capsule. These capsules are then used to compute collision costs with the surrounding world obstacles. 
We note that this cost is quick to implement---this can be done with only a few lines of Python.
An example output is shown in Figure~\ref{fig:trajopt}; see the supplemental video for more results.

\subsection{Motion Retargeting}

\algname{}'s flexibility is especially useful for motion retargeting, which often requires scaling or motion warping. 
Extensibility lets \algname{} tackle retargeting scenarios that require careful balance between motion fidelity and physical constraints. 
We demonstrate the versatility of \algname{} through several challenging retargeting scenarios, from full-body humanoid motion to hand-object interactions. 
Each task requires domain-specific costs, which can be prototyped quickly using auto-differentiated Jacobians. %
We implement sparse, keypoint‐based  retargeting inspired by~\cite{cheynel2023sparse}.
By aligning keypoint positions and preserving the relative distances and angular relationships between joints, our cost functions treat both body and hand cases identically. This unified approach allows our method to seamlessly adapt across diverse skeletal structures with varying scales and degrees of freedom. We discuss the details for each experiment below: %

\subsubsection{Full-body Humanoid Retargeting} \label{sec:humanoid-retargeting}
We transfer human motion onto the Unitree G1~\cite{unitree_g1} and H1~\cite{unitree_h1} humanoids, ensuring that it remains \textit{physically plausible within the scene}. Our inputs include human keypoint trajectories, the scene mesh, and information regarding whether each foot is in contact with the ground on a per-frame basis.
Motion transfer costs are designed such that they focus on using joint relationships to preserve the motion~\cite{cheynel2023sparse}, instead of keypoint positions which tend to bring the humanoid feet too close together~\cite{he2024learning}. For each pair of joints in the kinematic chain, we optimize their relative positions (scaled by learned per-link factors) and relative angles (using cosine similarity between joint vectors) to match input human keypoints. This allows the optimizer to handle the differences in limb proportions. We also add a cost to penalize the knee joints from becoming too close to each other. To ensure physical plausibility, we enforce floor contact constraints, foot orientation costs, self-collision avoidance, and joint limits. Example results are shown in Fig.~\ref{fig:retargeting_results}. The same cost ensures robust retargeting to robots with significantly different shapes (G1: 127cm, H1: 178cm); see the supplemental video for more details.

\subsubsection{Hand-Object Interaction Retargeting} \label{sec:hand-retargeting}
In this scenario, we use \algname{} to transfer human hand motions from DexYCB~\cite{chao:cvpr2021} to a robot hand. We use the same motion transfer cost used in full-body humanoid retargeting to align MANO~\cite{romero2017embodied} human hand motions to a robotic Shadow Hand~\cite{shadowHand}.
In parallel, we add a contact cost that maintains consistent contact between the robot hand and object surfaces. %
This modular architecture simplifies domain-specific objectives (e.g., contact), which improve results for this task. %

\section{Quantitative Results}

We evaluate \algname{} on modularity, flexibility, and computational efficiency. 
First, we show it can be used to implement the vector retargeting approach of Dex-Retargeting~\cite{qin2023anyteleop} to achieve comparable hand motion quality.
Second, we show how we can extend an IK implementation to simultaneously optimize the base pose of a Fetch mobile manipulator with its arm configuration.
Finally, we compare IK runtime performance, measuring how autodiff and analytical Jacobians scale against cuRobo~\cite{curobo_icra23} on CPU and GPU. All experiments were performed on a desktop PC with an AMD 5955WX CPU and NVIDIA RTX 4090 GPU.

\subsection{Reproducing Dex-Retargeting}

\begin{figure}
    \centering
    \includegraphics[width=0.7\linewidth, trim=3 3 3 3,clip]{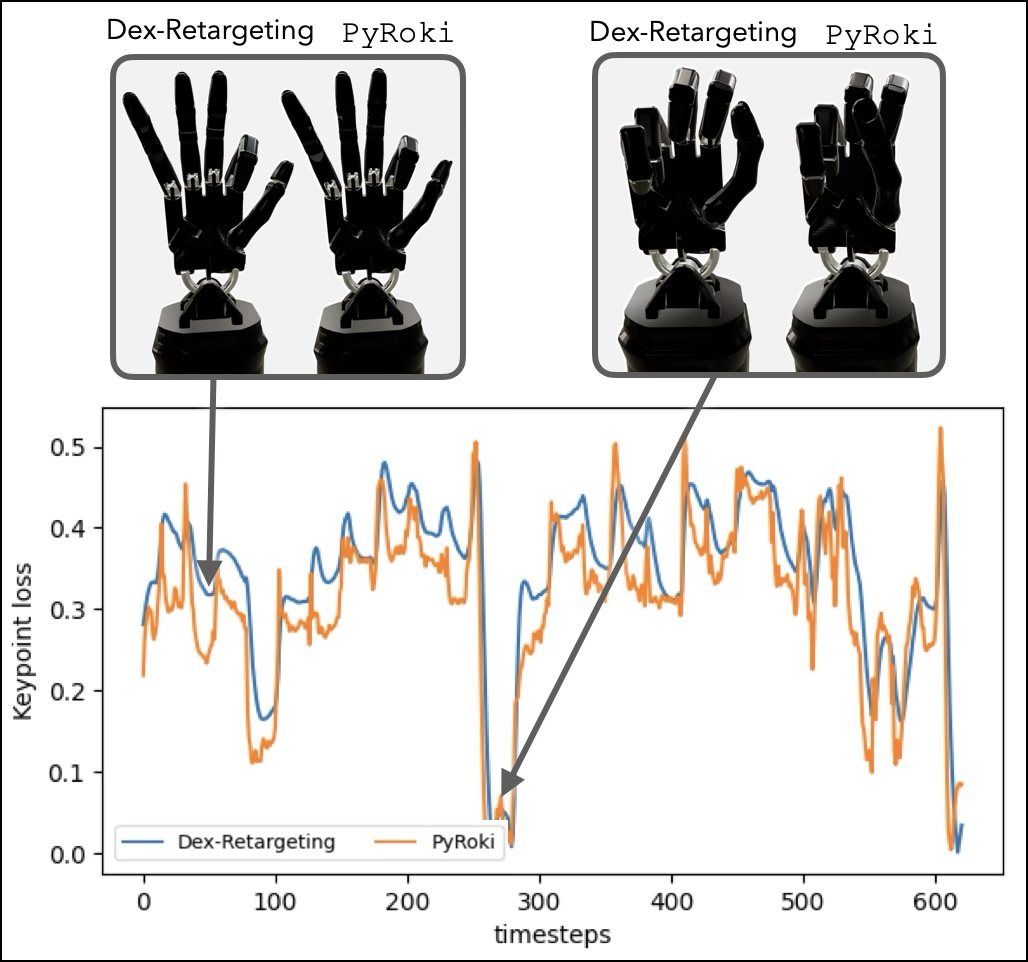}
    \caption{
        \textbf{Implementing vector-based hand retargeting.} 
        We implement the keypoint vector loss described in Dex-Retargeting~\cite{qin2023anyteleop} using \algname{}. Our global IK approach achieves slightly lower final costs than the original (diff. IK). %
    } \label{fig:dexretarget}
\end{figure}

To evaluate \algname{}’s modularity, we demonstrate its ability to replicate existing methods by re-implementing the vector retargeting cost in Dex-Retargeting~\cite{qin2023anyteleop}, a CPU-based differential IK optimization library.
We find that our implementation can produce visually similar motions that capture the same semantic hand poses, and reach very similar values for the keypoint cost (0.068 ± 0.070) for a human hand trajectory of 621 timesteps, as shown in Fig.~\ref{fig:dexretarget}. 
This problem can also be extended with new costs (e.g., contact loss in Sec.\ref{sec:hand-retargeting}) and batch processing on GPUs.

\subsection{Inverse Kinematics with a Mobile Base}
\label{sec:ik_mobile_base}
In mobile manipulation, solving inverse kinematics (IK) requires considering both the robot's arm and base positions. A common method is to first estimate a fixed base pose and then solve IK for the arm~\cite{chitta2012moveit,beeson2015trac}. However, this sequential approach can fail if the chosen base pose makes the target end-effector pose unreachable.
With \algname{}'s modular design, we can add the SE(2) base pose as an optimization variable and adjust pose cost such that the end effector's transform includes the base transform ($\mathbf{T}_{\text{world} \leftarrow \text{base}} \mathbf{T}_{\text{base} \leftarrow i}$). To evaluate this method, we sample reachable end-effector poses from a fixed base position and apply random translations to create target poses that may require base movement. An IK solution is deemed successful if the position error is under 5mm and the rotation error is below 0.05 radians.

As shown in Table~\ref{tab:mobile_base_comparison}, optimizing the base pose dramatically improves performance. With a fixed base, the solver achieves only a 15\% success rate with large position errors ($>$60cm). In contrast, simultaneous optimization of the base pose achieves a 100\% success rate with mean position errors under 0.005mm. This improvement highlights the practical utility of \algname{}'s extensibility. %

\begin{table}[t]
    \centering
    \begin{tabular}{lccc}
    \toprule
    \multirow{2}{*}[-0.75ex]{Mobile Base} & \multicolumn{2}{c}{Error (mean ± std)} & \multirow{2}{*}[-0.75ex]{\makecell{Success\\Rate (\%)}} \\
    \cmidrule(lr){2-3}
    & Pos. (m) & Rot. (rad) & \\
    \midrule
    Static & 0.616 ± 0.600 & 0.470 ± 0.426 & 24 \\
    Optimized & \textbf{5.40e-6 ± 4.31e-5} & \textbf{0.0 ± 0.0} & \textbf{100} \\
    \bottomrule
    \end{tabular}
    \caption{\textbf{Optimizing base pose for IK with mobile robot}: 
    The inverse kinematics task in Sec.~\ref{sec:ik_trajopt} can be modified to incorporate the base pose of the Fetch robot as an optimizable variable, which significantly improves both position and rotation accuracy.}
    \label{tab:mobile_base_comparison}
\end{table}

\begin{table}
    \centering
\sisetup{table-number-alignment = center, detect-weight, detect-inline-weight=math}
\begin{tabular}{l S[table-format=4.1] S[table-format=6.1] S[table-format=4.1] S[table-format=6.1]}
\toprule
\multirow{2}{*}[-.75ex]{Batch} & \multicolumn{2}{c}{\texttt{PyRoki} (CPU)} & \multicolumn{2}{c}{\texttt{PyRoki} (GPU)} \\
\cmidrule(lr){2-3} \cmidrule(lr){4-5}
& {Analytical} & {AutoDiff} & {Analytical} & {AutoDiff} \\
\midrule
1    & 5.9 & 36.9   & \bfseries 3.6 & 11.4 \\
10   & 48.2 & 332.4 & \bfseries 3.6 & 13.1 \\
100  & 396.7 & 2173.0 & \bfseries 4.7 & 26.6 \\
1000 & 3037.6 & 19949.0 & \bfseries 15.5 & 253.1 \\
2000 & 5534.7 & 39284.8 & \bfseries 32.8 & 624.4 \\
\bottomrule
\end{tabular}
    \caption{\textbf{IK-Beam runtime for analytical vs. autodiff Jacobians.} We compare inverse kinematics solve times (ms) on CPU and GPU. Analytical Jacobians provide significant speedup, especially at higher batch sizes.
    IK-Beam uses 64 initial seeds; a batch size of 1000 means that up to 64000 LM steps are computed in parallel.
    }
    \vspace{-1em}
    \label{tab:analytical_autodiff}
\end{table}

\begin{table*}[h]
  \centering
  \small
  \sisetup{table-format=2.2,scientific-notation=true}
  \begin{tabular}{
      c
      c c S[table-format=1.2e1] S[table-format=1.2e1]  %
      c c S[table-format=1.2e1] S[table-format=1.2e1]  %
  }
    \toprule
    \multirow{2}{*}{Batch} &
    \multicolumn{4}{c}{cuRobo~\cite{curobo_icra23}} &
    \multicolumn{4}{c}{IK-Beam (\algname{})} \\
    \cmidrule(lr){2-5}\cmidrule(lr){6-9}
     & {Time (ms)} & {Succ.\,(\%)} & {Pos.\,Err} & {Ori.\,Err (rad)}
     & {Time (ms)} & {Succ.\,(\%)} & {Pos.\,Err\,(mm)} & {Ori.\,Err (rad)} \\
    \midrule
     1    & 5.03  & 100 & 7.40e-4 & 2.55e-6  & 3.58  & 100 & 8.50e-5 & 5.65e-7 \\
     10   & 5.27  & 100 & 2.67e-3 & 5.58e-6  & 3.60  & 100 & 1.65e-4 & 2.72e-7 \\
     100  & 6.76  & 100 & 3.59e-3 & 7.05e-6  & 4.70  & 100 & 2.99e-4 & 4.19e-7 \\
     1000 & 26.37 & 100 & 4.70e-3 & 6.96e-6  & 15.54 & 100 & 2.55e-4 & 3.89e-7 \\
     2000 & 50.32 & 99.95 & 4.31e-3 & 6.93e-6 & 32.81 & 100 & 3.04e-4 & 4.67e-7 \\
    \bottomrule
  \end{tabular}
  \caption{
        \textbf{Comparison between cuRobo and IK-Beam.}
        IK-Beam, which is implemented using \algname{}, is faster and converges to lower errors.
        Experiments are run on the Franka Panda robot with a modified version of the official cuRobo~\cite{curobo_icra23} benchmarking script.
        Following cuRobo, reported errors are the 98th percentile of all solutions in the batch.
  }
  \vspace{-0.5em}
  \label{tab:curobo_vs_pyroki}
\end{table*}

\subsection{Runtime Analysis} %
\algname{} supports both analytical and autodiff Jacobians, parallelized optimization, and different computation platforms (CPU, GPU, TPU).
To explore the advantages of this flexibility, we explore IK for the Franka Panda robot using a beam search-inspired algorithm that we call \textit{IK-Beam}.
IK-Beam is designed to be both robust and simple to implement using \algname{}.
Given a target pose, IK-Beam begins with 64 start seeds.
We run 16 total Levenberg-Marquardt steps.
The first 6 LM steps optimize all 64 initializations in parallel.
We then discard all but the 4 seeds with lowest error, which are optimized for the 10 final LM steps.
The returned solution is the best from these 4 seeds.

Table~\ref{tab:analytical_autodiff} presents IK-Beam solve times across varying batch sizes, Jacobian options, and CPU/GPU compute platforms. We find that analytical Jacobians provide speedups ranging from 3x-19x over autodiff.
GPU parallelization helps runtime scale better with increasing batch sizes.

We additionally compare with cuRobo~\cite{curobo_icra23}, a popular GPU-accelerated IK library (Table~\ref{tab:framework_comparison}).
Depending on batch size, we find that IK-Beam is between 1.4x and 1.7x faster, while converging to lower errors.
We believe that the accuracy improvement is mostly explained by the optimizer---cuRobo uses L-BFGS, while \algname{} uses LM (Sec.~\ref{sec:solver}).

\section{Conclusion}

We present \algname{}, a modular, extensible, and cross-platform toolkit for robot kinematic optimization.
A variety of robot kinematic optimization problems can be implemented and extended simply by composing variables and cost functions, while running efficiently on CPUs, GPUs, and TPUs. For example, we can reuse the same collision costs throughout all the tasks presented in the paper---from inverse kinematics, trajectory optimization, to motion retargeting with humanoids. 
\algname{} is open-source and invites users to extend its capabilities.

\section{Acknowledgement}
This project was funded in part by NSF:CNS-2235013, IARPA DOI/IBC No. 140D0423C0035, and DARPA No. HR001123C0021; Chung Min Kim and Brent Yi
are supported by the NSF Research Fellowship Program, Grant DGE 2146752. We thank Kevin Zakka and Justin Kerr for fruitful technical conversations.

\renewcommand*{\bibfont}{\footnotesize}
\printbibliography
\end{document}